\documentclass[10.5pt]{article}
\usepackage[utf8]{inputenc}
\usepackage{graphicx}
\usepackage{xcolor}
\usepackage[fleqn]{amsmath}
\usepackage{algpseudocode}
\usepackage[boxruled,linesnumbered]{algorithm2e}
\usepackage{upgreek}
\usepackage{graphicx}
\usepackage{soul}
\usepackage[preprint, nonatbib]{neurips_2019}
\usepackage{url}

\title{
{Neural Machine Translation with Monte-Carlo Tree Search}}
\author{%
  Jerrod Parker \\
  Department of Computer Science \\
  University of Toronto \\
  \texttt{jparker@cs.toronto.edu} \\
  \And
  Jerry Zikun Chen \\
  Department of Computer Science\\
  University of Toronto\\
  \texttt{jzchen@cs.toronto.edu} \\
}

\begin{document}
\maketitle
\begin{abstract}
Recent algorithms in machine translation have included a value network to assist the policy network when deciding which word to output at each step of the translation. The addition of a value network helps the algorithm perform better on evaluation metrics like the BLEU score. After training the policy and value networks in a supervised setting, the policy and value networks can be jointly improved through common actor-critic methods. The main idea of our project is to instead leverage Monte-Carlo Tree Search (MCTS) to search for good output words with guidance from a combined policy and value network architecture in a similar fashion as AlphaZero \cite{silver2017mastering_5}. This network serves both as a local and a global look-ahead reference that uses the result of the search to improve itself. Experiments using the IWLST14 German to English translation dataset show that our method outperforms the actor-critic methods used in recent machine translation papers. Full code for this
work is available at \url{https://github.com/chenziku/NMT-MCTS}.

\end{abstract}

\section{Introduction}

Many algorithms in machine translation use an encoder-decoder model, where input sentences are encoded into feature vectors. Translated sentences are generated word by word from the decoder. For each word output, there is a conditional distribution over the word given the input sentence and previous output words. These models are often trained by conditioning on the true output so far in an attempt to maximize the probability of the correct next word. For example, if we have output $(t-1)$ words and an input sentence $(x_1, \cdots, x_m)$ of length $m$, our distribution over the next output is $p(y_t | y_{t-1}, \cdots, y_1; x_1, \cdots, x_m)$. This can often choose words that are good in the immediate future rather than optimal sentences in the long run. 

Despite this limitation, the current state-of-the-art in neural machine translation such as \cite{SOTA1} only uses a policy network. In the past, algorithms in machine translation \cite{Actor-critic-sequence_1, nmt_value_2} as well as image captioning \cite{DRL_captioning_4} managed to improve upon policy-only methods by adding a value network. This implies that the current state-of-the-art models can likely be improved by the addition of a value network, which can be trained jointly with the policy. The policy network is our conditional language model, such as a transformer \cite{transformerPaper} that outputs a distribution of next words given what we have seen at each time step. The value network predicts the expected reward, the BLEU score, that we would obtain given the current output if we continue following the policy to completion of the sentence. One benefit of jointly training the policy and value networks is that it helps guide the policy network to learn to optimize for longer term rewards such as the final BLEU score. Furthermore, recent papers \cite{Actor-critic-sequence_1, DRL_captioning_4} have used actor-critic methods (see Appendix 8.1) to jointly improve the policy and value networks.

A problem with these algorithms is that they produce unstable training and require tweaks such as a target critic network \cite{Actor-critic-sequence_1} or curriculum learning \cite{DRL_captioning_4} to converge on a quality model. This is likely due to the high variance gradient estimates that are used in these long sequence prediction tasks such as machine translation where the target sentence can be over 60 tokens. 

We improve on several existing reinforcement learning methods in neural machine translation \cite{Actor-critic-sequence_1,nmt_rl_study_3} by using Monte Carlo Tree Search (MCTS) in a way similar to AlphaZero \cite{silver2017mastering_5} so that our model has more stable updates. MCTS in AlphaZero achieved excellent performance in Go where long sequences of moves are predicted, which showed it's potential as a method to produce more accurate updates to the model than actor-critic methods such as those used in \cite{DRL_captioning_4}. We conduct several experiments to analyze where these benefits are derived from as well as discuss limitations and possible improvements to our current model.

\section{Related Work}
Previous works have leveraged a value network to complement the policy network in machine translation \cite{Actor-critic-sequence_1, nmt_value_2}, image captioning \cite{DRL_captioning_4} and playing Go \cite{silver2017mastering_5}. \cite{DRL_captioning_4} designs a model for image captioning and trains a policy and value network in a supervised manner using embedded rewards. The model is then updated through an actor-critic reinforcement learning method similar to A2C (see Appendix 8.1). The authors found that the global guidance introduced by the value network greatly improves performance over just using a policy network. Another paper which uses an actor-critic method in machine translation is \cite{Actor-critic-sequence_1} which trains their policy and value networks by reinforcement learning but allows the value network to also take the true output as its input. This method helps to improve the policy by allowing the policy to directly optimize for BLEU score. \cite{nmt_value_2} trains a value network to assist the policy in the decoding stage of machine translation but does not do any joint training of the value and policy networks. 

As a significant milestone in reinforcement learning, paper \cite{silver2017mastering_5} uses self-play and MCTS to train a policy and value network with a shared body, which led to efficient optimization for move predictions in the game of Go. The MCTS algorithm was shown to be a powerful policy improvement and policy evaluation method.

One of the big differences of our method compared to previous works is the addition of a value network with reinforcement learning methods to jointly update the policy and value networks for neural machine translation. The only papers we found that did this were in \cite{Actor-critic-sequence_1}, and an image captioning paper \cite{DRL_captioning_4} where the authors used an actor-critic methods similar to A2C. In contrast, we used MCTS which we found to have some advantages over using the method from \cite{DRL_captioning_4} in neural machine translation which we expand upon in section 5.2. In \cite{silver2017mastering_5}, MCTS is applied to the game of Go instead of machine translation, which we are interested in. One of our contributions is showing that MCTS can be used successfully in this domain by outperforming a model trained using the actor-critic method from \cite{DRL_captioning_4}. Another contribution is our investigation into where the benefits of the MCTS are derived from (section 5.2) which will help guide future research.  

\newpage
\section{Diagrams}
\begin{figure}[!ht]
    \includegraphics[width=16cm]{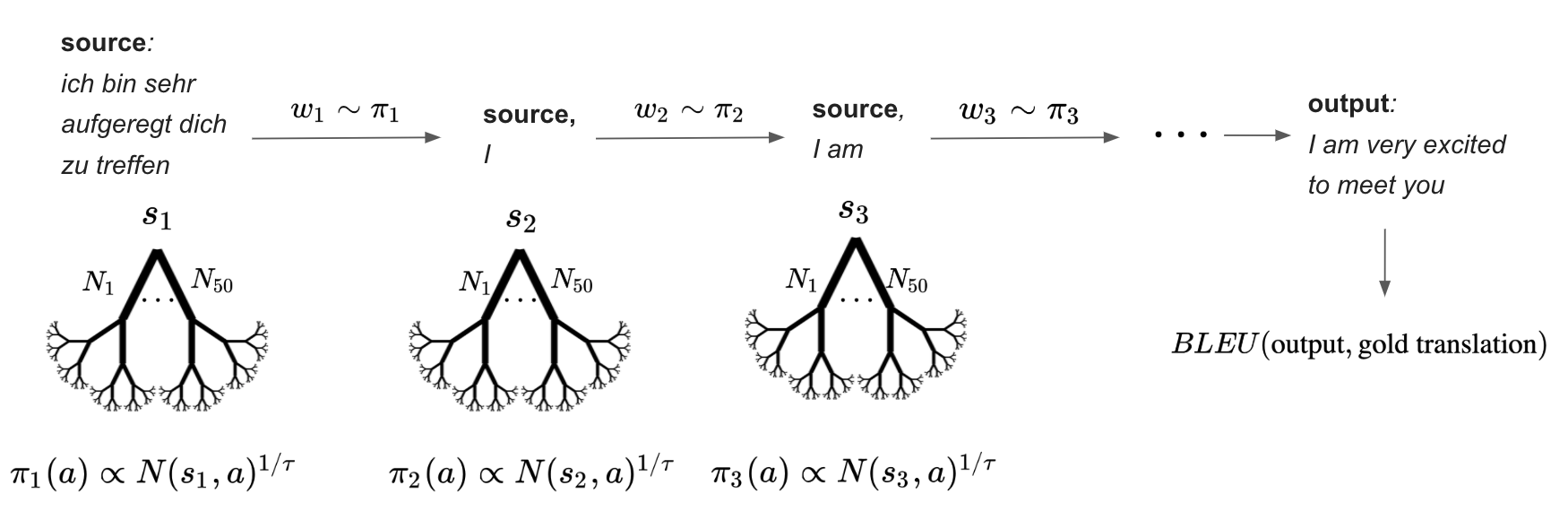}
    \caption{Process of translating one sentence. Note that these two figures are modifications of the figures in \cite{silver2017mastering_5}. At each time step we run $Y$ simulations each following from figure 2. After the $Y$ simulations we end up with visitation frequencies $N_i$ for each child $i$ of the root node. We then generate our next action at time step $j$ from the distribution $\pi$ which has probability of a particular action $i$, 
    ${\pi}_i \propto N(s_j, a_i)^{1/\uptau}$ where $s_j$ is the state composed of the source sentence and the output so far. $\uptau$ is a chosen temperature hyper-parameter that controls exploration. The translation finishes when $w_j$ is $EOS$ (End Of Sentence token).}
    \label{fig:translation}
\end{figure}

\begin{figure}[!ht]
    \includegraphics[width=14cm]{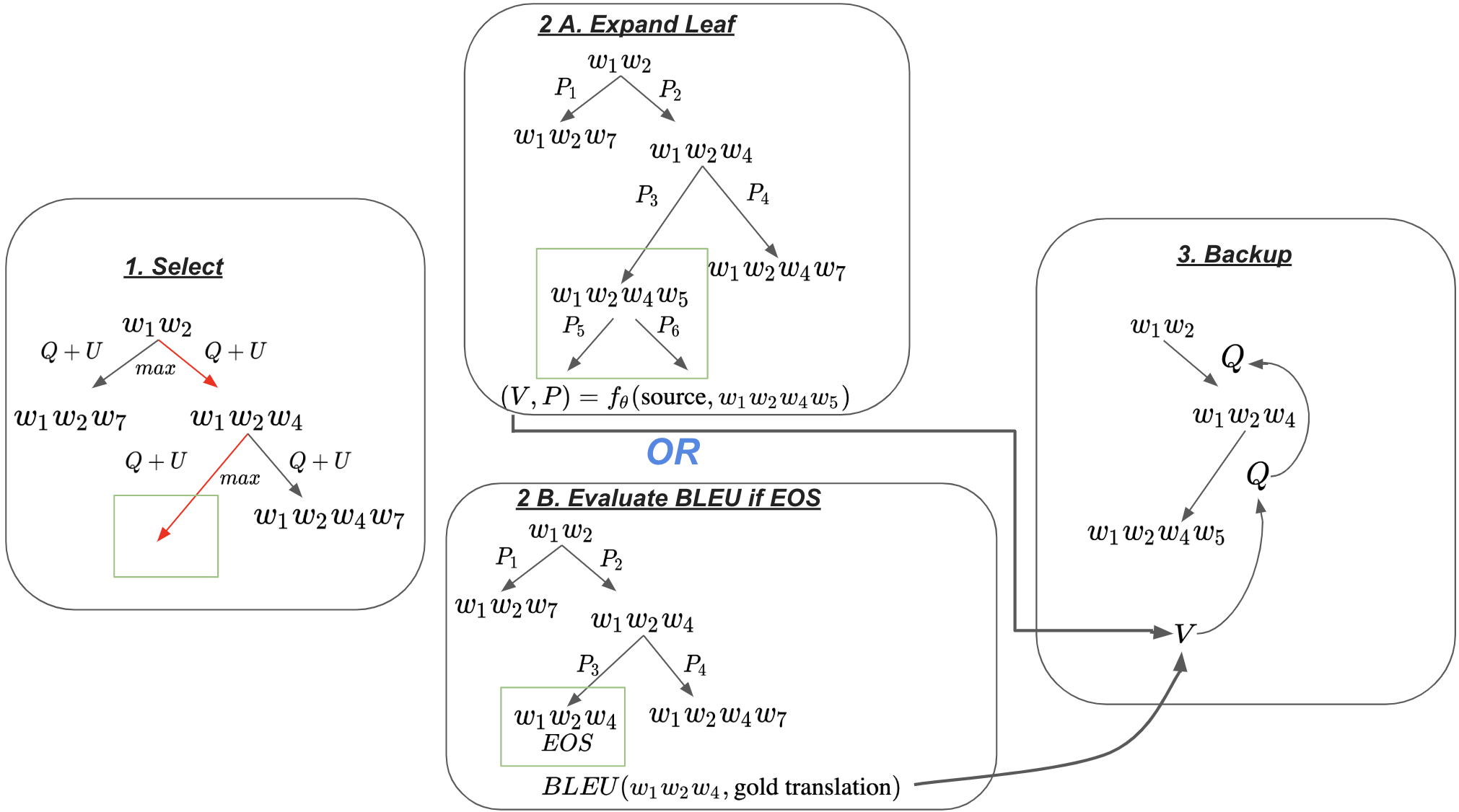}
    \caption{Corresponds to algorithm 1 below. When choosing our next word, we run $Y$ simulations, each of which consists of first, continually selecting branches to move to based on the branches' $Q+U$ value (see algorithm 1) until we reach a branch with no child. Second, there are two possibilities at this step. A) If the leaf node corresponds to a full translation (contains EOS) then we calculate the BLEU score of our current translation vs. the true translation as the value for the node. B) We expand this leaf node by running the neural network $f_\theta$ with the translation up until this node as well as the input sentence. This returns ($P$,$V$) where $P$ is the prior probabilities of taking each action from this state, and $V$ is the model's value estimate at this state. Lastly, we back up the value computed at the second step by updating the $Q$ values for each edge along the simulated path. The $Q$ value is updated to be the average of all values ever backed up along that edge. 
    }
    \label{fig:rollout}
\end{figure}

\newpage
\section{Formal Description}
\textbf{Tree node attributes} Each node contains a dictionary called 'Edges' which maps an action to an edge which has attributes, $N$: number of times the edge has been traversed, $W$: sum of values backed up to this edge over all simulations, $Q: W/N$, $P$ which is prior probability of taking this edge which is assigned by the network during the expansion of this node. Each node also has a 'children' dictionary which maps an action to a child tree node. The node also contains the translation from the root to this node. The attributes $W$ and $N$ of each edge are initialized as 0. 
\begin{algorithm}[H]
 \SetAlgoLined
 Input: root: is root node of current tree\\
 hyper-parameters: $0 < \uptau$, $0 < c_{puct}$, numSimulations\\
 let an edge's U value = $c_{puct}*P\frac{\sqrt{N_{parent}}}{N}$ \\
 \For{$sim\gets 1, numSimulations$}{
    curNode = root //we have root.parent==NULL\\
    \While{True}{
      $N_{parent}$ = curNode.parent.edges[curNode.parentEdgeUsed].N //unless root\\
      a = edge from curNode with max Q+U\\
      curNode.edges[a].N += 1\\
      \eIf{$curNode.children[a]==Null$}{
        Break
       }{
        child = curNode.children[a]\\
        child.parent = curNode, child.parentEdgeUsed = a\\
        curNode = child
      }
    }
    // now have reached leaf node. If terminal state(hit EOS) then don't expand\\
    // want to backup the true BLEU score during training when at EOS \\
    \eIf{curNode is terminal}{
        value = BLEU score of translation to this node
    }
    {
        (priors,value) = network(curNode.state) \\
        \hl{curNode.children is initialized with top 50 most probable actions from priors}
    }
    // now update Q-values of edges along path taken (doing backup)\\
    \While{curNode.parent not Null}{
        parent = curNode.parent, edgeUsed = curNode.parentEdgeUsed \\ 
        parent.edges[edgeUsed].W += value\\
        $parent.edges[edgeUsed].Q = parent.edges[edgeUsed].W / parent.edges[edgeUsed].N$\\
        curNode = parent
    }
 }
 sumPriors = $\sum_{a} {root.edges[a].P}$\\
 sumVisitations = $\sum_{a} {root.edges[a].N}^{\frac{1}{\uptau}}$ \\
 //simVisitationProbs is dict of wordIndice to probability found in our MCTS\\
 \For{a in root.edges.keys()}{
    \hl{simVisitationProbs[$a$] = $\frac{{root.edges[a].N}^{\frac{1}{\uptau}}}{sumVisitations} * sumPriors$}
 }
 \Return{simVisitationProbs}
 
 \caption{RunSimulations}
\end{algorithm}

\noindent
\textbf{Key differences from AlphaZero} Firstly, there aren't two players so each simulation is only from one perspective. Secondly, at each node we only store the top 50 edges ranked by their prior probability from our network. This means that when we get our visitation frequencies among those edges, we don't want to update our policy network directly on the probabilities proportional to the visitation frequencies. This is because these probabilities add to 1 but during the simulations we never gave the algorithm a possibility of visiting the other 6450 branches. To compensate for this, we multiply each probability in 'simVisitationProbs' by the sum of the prior probabilities of the 50 edges leaving that node and then only calculate the cross entropy loss on these 50 actions we used. 
We presume this does not drastically change the probabilities of the actions that we never gave the tree search the chance to visit. The lines of the pseudo-code that correspond to these differences are highlighted in yellow. 

\begin{minipage}{1\linewidth}

\begin{algorithm}[H]
 \SetAlgoLined
 INPUT: sentences which is list of (germanSentence,englishSentence) pairs\\
 trainingLists []\\
 \For{(input,targetOutput) in sentences}{
    root = init node with current translation of BOS token, run policy to get priors.\\
    tmpLists = empty list\\
    currentTranslation = empty list\\
    \While{root isn't terminal}{
        simVisitationProbs = run algorithm 1 with root node\\
        nextWord = sample from normalized simVisitationProbs \\
        tmpLists.append([simVisitationProbs,root.state])\\ // root.state contains both input sentence and current translation\\
        root = root.children[nextWord]\\
        root.parent = NULL \\
        currentTranslation.append(nextWord)
    }
    bleuScore = Bleu(currentTranslation,targetOutput)\\
    append blueScore to each list in tmpLists\\
    append all lists in tmpLists to trainingLists\\
 }
 \Return{trainingLists}
 \caption{SimSentences}
\end{algorithm}
 \end{minipage}

\begin{minipage}{1\linewidth}
\begin{algorithm}[H]
 \SetAlgoLined
 INPUT: run Algorithm 2 and get trainingLists\\
 sumLoss = 0\\
 \For{(simProbs,state,bleuScore) in trainingLists}{
    policyProbs,value = network(state) \\ 
    \hl{// only want cross entropy on words in simProbs (which are the keys)}  \\
    \hl{filteredProbs = [policyProbs[i] for i in simProbs.keys()]} \\
    loss = ${(bleuScore-value)}^2 + simProbs*log(filteredProbs) + c*{\theta}^2$ \\ 
    sumLoss += loss  
 }
 backpropogate on sumLoss\\
 gradient descent on parameters of network $\theta$\\
 \caption{UpdateNetwork}
\end{algorithm}
 \end{minipage}

\section{Experiments}
In this section, we conduct experiments to gauge the effectiveness of using MCTS as a policy improvement operator in neural machine translation as well as to explore where the benefits of the MCTS stem from.

\subsection{Dataset, Implementation, and Results}
\textbf{Dataset} The dataset we use for our experiments is the IWSLT14 German to English translation dataset which had training size of 149184, validation size of 6784, and test size of 6400 sentences. Our evaluation metric was BLEU calculated by SacreBLEU \cite{sacrebleu}, which is one of the main metrics used to compare machine translation models. The tokenization is done by FairSeq which has a script for preprocessing this dataset.

\textbf{Implementation}
We used the transformer architecture for the policy network from \cite{transformerPaper} except reduced the size to allow for faster computation and training. We used layer sizes as follows: word embedding dimension is 128, number of attention heads is 8, number of encoder and decoder layers is 4, the dimension of the feedforward network is 512. The value network uses the same architecture except that the output layer mapped to size 1 instead of the length of the target vocabulary.
The supervised training of the policy was done using cross entropy loss with teacher forcing, and the Adam optimizer with learning rate schedule from \cite{transformerPaper}. This model is referred to below as 'Supervised Policy'. The initial training of the value network was done in the same manner as in \cite{DRL_captioning_4} which we describe in the appendix. We used a dropout of 0.2 in the encoder and decoder layers of our models during training. 
As one of our baselines, we used the 'Supervised Policy' updated with policy gradients estimated through REINFORCE as done in \cite{nmt_rl_study_3} which achieved SOTA performance in 2018. We refer to this model as 'Policy+RL'. We then jointly trained the supervised policy and value network using an algorithm similar to A2C as done in \cite{DRL_captioning_4}. We refer to this model as 'Actor-Critic'. Further details on the training of 'Policy+RL' and 'Actor-Critic' are in appendix 8.1. 
Finally we also jointly trained the policy and value network using MCTS as described in algorithms 1-3. Further specifics of the implementation are in Appendix 8.2. The best hyperparameters we found were $c_{puct}=0.5$,$\uptau = 1$ and 100 rollouts per action. Note that because of computational constraints the number of rollouts per action wasn't increased past 100. We would expect improved performance if this number was increased as it would lead to lower variance updates to the model.

\textbf{Results}
We compare the results of the models below. For each model, after training we take the policy from that model and run it on the test set using greedy decoding to get the SacreBLEU \cite{sacrebleu} score. This is because of the large amount of time it would take to run our MCTS on the entire test set.

\begin{center}
\begin{tabular}{l*{3}{c}r}
\textbf{Methodology}
& \textbf{BLEU}  \\
\hline
Supervised Policy
&  22.32 \\
MCTS
&  27.29 \\
Actor-Critic   
&  26.95 \\
Policy+RL  
&  26.96 \\
\end{tabular}\\
\end{center}
\noindent
An observation from our experiments is that we weren't actually able to improve the policy using the actor-critic method in \cite{DRL_captioning_4}. Several tweaks had to be made to the actor-critic algorithms in \cite{Actor-critic-sequence_1,DRL_captioning_4} such as having target networks updated only so often, as well as curriculum training to allow the algorithm to converge on a quality model. We didn't implement these additions which could be the cause of our actor-critic model not improving past the 'Policy+RL' model. 

\subsection{Where do the benefits of MCTS come from?}
 We now explore where the benefits of MCTS over these baseline models come from. There are two clear areas where the benefits of our methods could be derived from that we will explore. 
    \begin{enumerate} 
        \item \textbf{Is it just because we’re doing so many simulations at each step that we could get the actor-critic model to do similarly by increasing batch size? In other words, are we getting a lower variance gradient estimate purely from increased computation?} 
        
        To perform a fair comparison, we increased the batch size for the policy gradient method until the run time to process the batch was equivalent to the time per batch for our MCTS, which was variable but around 22 minutes. For MCTS this corresponds to 256 sentences and for the policy gradient method this corresponds to 27200 sentences. This is a rough comparison as we could speed up the MCTS significantly as mentioned in the next section. 
        Even when increasing the batch size to 27200 sentences, the validation BLEU score for 'Policy+RL' was 27.79. This BLEU score is much lower than the BLEU of 28.37 obtained from training the policy using the MCTS method. This shows that the advantage of the MCTS over the 'Policy + RL' model likely does not only come from the increased computation. 
        
        \item \textbf{Is it the addition of the value network that is improving the performance over the 'Policy + RL' model?}
        
        To test this, we adjusted the MCTS algorithm to run without a value network, where we only backed up values from states which corresponded to a translation ending in an $EOS$ token. The value backed up was the BLEU score of the translation. It was only during these backups that the visitation frequencies of the edges on the path were incremented. This was fair since we did the same thing during training when a value network is present. Results of the policy obtained from using this modified algorithm during training are in Appendix 8.3, where it achieves better performance than ‘Policy + RL’ (BLEU of 28.26 vs 27.78) and similar performance to the version of MCTS that uses a value network (BLEU of 28.26 vs 28.37). We hypothesize that the reason for this improvement is that the MCTS algorithm only updates one state at a time and the visitation frequencies that are used to update the policy were guided by both accurate Q-values as well as our previous prior probabilities. This leads to low variance gradients, whereas our policy gradients estimated with REINFORCE are trying to update the probabilities of many states using only a sparse reward for the final state. These gradient estimates have high variance since many of the sentences contain more than 40 tokens, which makes it hard for the algorithm to determine the effect each word in the sequence had on the reward.
    \end{enumerate}

\section{Limitations and Future Improvement}

\begin{enumerate}
    \item \textbf{Computational Cost} Running this algorithm takes a lot of computational resources and time. For example, translations for 16 sentences of length 19 took approximately 2.5 minutes when only using 100 simulations per step. This means it would be quite time consuming to train a policy and value network from scratch with our implementation. Instead, what we found is that using a pre-trained policy and value network, we can quickly improve these networks within a couple of gradient steps, which makes it computationally feasible. The main change that we think would improve the speed of our algorithm is to use more efficient inter-process communication. Currently, the majority of the time is spent on processes waiting for results from the model. For example, each time a node in a tree needs to be expanded, the process in charge of that tree sends the current state to a process which runs the neural network with that state and then returns the results. Currently the time it takes for the model to run is 10 times less than the amount of time that the process, which sends a request to the model, waits for results.
     
    \item \textbf{Search Diversity} In the case of machine translation, there were 6565 tokens we could output at any one step. This large action space would cause the memory storage of each node to be very large (each array would be size 6565) so we only stored a small subset of these actions as children of a tree node. In our case, since the model was initially given pre-trained policy and value networks, the policy had already decently narrowed down what the good actions were at each step by the probability mass that it assigned to each word. This allowed for us to only use the top 50 children of each state based on the prior policy probabilities. This is not a perfect solution as at times the true word will be outside the top 50 predicted by the policy but our algorithm relies on this not happening too often. Our current implementation would make it impossible to train from scratch without some modifications since we are throwing out the majority of the possibilities at each step. Even with modifications, it would be difficult to train from scratch because of the massive action space which would be hard to fully explore. It might be possible to utilize a computational oracle constructed from the ground truth to better select the candidate words at each tree expansion.
    
    \item \textbf{Reinforcement Learning} In the training phase, we have access to the ground truth translation tokens at each time step. The gold translations are not utilized until we simulate an entire sentence to compute BLEU scores. This can cause undesirable paths to be taken during the MCTS while the model is training. There is potential to utilize imitation learning techniques that interpolates between pure supervision and reinforcement learning to better guide MCTS training of the policy along simulation paths.
\end{enumerate}

\section{Conclusion}
In this project, we presented a modification of the MCTS algorithm from AlphaZero \cite{silver2017mastering_5} to jointly train policy and value networks in neural machine translation. We compared the performance of this algorithm with a model which was updated using estimated policy gradients as well as a model that updates  value and policy networks with an actor-critic method similar to A2C. Our results on the IWSLT14 dataset showed performance increases in test set BLEU scores over the other two methods. This shows that many of the current SOTA models such as \cite{SOTA1} which purely use a policy could likely be improved through updating the policy using our MCTS algorithm. This could either be done by using our modified MCTS which only uses a policy or our original MCTS in which case a value network must be pre-trained for that policy. This algorithm looks to act as a promising alternative to other actor-critic methods such as those used in \cite{Actor-critic-sequence_1, nmt_value_2}. Lastly, our experiments were confined to the domain of language translation but it is possible to apply this method to other sequence prediction tasks such as in image captioning which is a closely related to the problem of neural machine translation.

\newpage
\bibliographystyle{IEEEtranS}
\bibliography{dissertationbib}
\newpage

\section {Appendix}
\subsection{Formulas}
In the model above called 'Policy + RL', the policy is updated using estimates of the policy gradient which are obtained through REINFORCE \cite{Williams1992}. To implement this, we first simulate a translation to the input sentence of length $T$ which gives us states $s_1 ... s_T$ during the translation and final BLEU score b. Then, we suppose parameters of the policy are $\pi$. The gradient estimate for the policy is then estimated as $\sum_{t=1}^{T} {\nabla}_{\pi} b*log p_{\pi}(a_t | s_t)$ 

In the model 'Actor-Critic', the policy and value network are jointly updated as in \cite{DRL_captioning_4}. The algorithm works as follows: 1. We simulate a translation to the input sentence of length T which gives us states $s_1 \cdots  s_T$ during the translation and final Bleu score $b$. 2. Suppose parameters of the policy are $\pi$ and parameters of the value network are $\theta$. The gradient estimate for the policy is then estimated as: $\sum_{t=1}^{T} {\nabla}_{\pi} log p_{\pi}(a_t | s_t)(b-v_{\theta}(s_t)$ and the gradient estimate for the value network is: $\sum_{t=1}^{T} {\nabla}_{\theta} v_{\theta} (s_t)(b-v_{\theta}(s_t))$.

\subsection{Additional Implementation Details}

\textbf{Initial Value Network Training} To initially train the value network, we used the same training scheme as in \cite{DRL_captioning_4} where for each input sentence, a translation is simulated $x_1\cdots x_T$ using the trained policy (and greedy decoding) and a BLEU score $b$ is produced. Then, an index $j$ is uniformly picked within this translation and we use $x_1 \cdots x_j$ as input to the value network decoder, where $(v(x_1\cdots x_j)-b)^2$ is minimized. Not every prefix of x is used because there is a strong correlation between successive prefixes (the state corresponding to that prefix). So we do not want to update the model using similar states successively as this could lead to over-fitting.

\textbf{MCTS} We processed batches of 64 sentences at a time and created one process per sentence in our batch so that the tree searches for each sentence could run in parallel. There is one main process that controls the model. Each time a roll-out in one of the trees reaches a leaf node and needs to expand a state, it sends the state tensor to the main process. Thus, the neural network is able to run with a batch of state tensors and then sends the expansion results back to each of these processes. The inter-process communication took the majority of the time, as we were able to write the code for the rollouts in optimized Cython which sped up the algorithm drastically. 

Now from each of these sentences, training data were gathered (input sentence, current translation, $\pi$, BLEU) as described in the algorithm section as well as in Figure 1. To update the model we would first process 4 batches (256 sentences), and then perform 8 iterations of randomly drawing 256 samples from the dataset collected during the processing of these batches, run the model and get the loss as described in Algorithm 3, back propagate, then we do a gradient step. 

\subsection{Is a Shared Architecture Necessary? Is the Value Network Necessary?}
We did several experiments to see if a shared architecture for the policy and value network would lead to performance benefits like it did in \cite{silver2017mastering_5}. To do this, we used the policy encoder as the encoder for both the policy and value networks. We hypothesized that the encoder's job for both the value and policy networks are similar in nature which means that the shared architecture can help performance. The encoder learns to create embedding of the input sentences which can be useful for learning what the long term value of choosing the next word is as well as what the probability of choosing that next word should be. To test this theory, we did several experiments where everything was kept constant except whether the encoder was shared or not.

Another interesting question we investigated was if the value network was even necessary. We mention how we modified the MCTS algorithm to only use a policy in section 6. This model is 'No Value' in the table below.

To investigate both of these questions, we do several experiments where the hyper parameter $c_{puct}$ is manipulated while temperature $\tau=1$, number of rollouts per action$=100$, Adam learning rate $=1e-4$. The runs took close to 5 hours per set of hyper parameters so we only tested several. Here are the validation set BLEU scores of our experiments:

\newpage
\begin{table}[ht]
\caption{BLEU Score Results} 
\centering 
\begin{tabular}{c c c c c}
\hline\hline 
$c_{puct}$ & \textbf{Joint Network} & \textbf{Disjoint Network} & \textbf{No Value}  \\ [0.5ex]
\hline 
0.1 & 28.16 & 28.37 & 28.07 \\ 
0.5 & 28.29 & 28.34 & 28.18 \\
1 & 28.20 & 28.11 & 28.26 \\
5 &  &  & 27.82 \\ [1ex]
\hline 
\end{tabular}
\label{table:n}
\end{table}

$$\textbf{Policy + RL} \text{ produces a BLEU score of } 27.78$$

\noindent
The results for shared vs disjoint architectures look very similar and will require more extensive experiments to determine if there is a true difference. 
The results from running the MCTS algorithm with only a policy look promising as well and look to have just slightly worse performance than when we include a value network. These results show that there may be some benefit in including the value network but would also require more extensive experimentation to draw firm conclusions. 

\end{document}